\documentclass[lettersize,journal]{IEEEtran}

\usepackage{cite}
\usepackage{amsmath,amssymb,amsfonts}
\usepackage{graphicx}
\usepackage{textcomp}
\usepackage{algorithm}
\usepackage{algpseudocode}
\usepackage{array}
\usepackage{multirow}
\usepackage{booktabs}
\usepackage{caption}
\usepackage{bm}
\usepackage{nth}
\usepackage{bbm}
\usepackage{graphicx}
\usepackage{wrapfig}
\usepackage{lipsum} 
\makeatletter
\let\NAT@parse\undefined
\makeatother
\usepackage{hyperref}
\hypersetup{hypertex=true,
colorlinks=true,
linkcolor=blue,
anchorcolor=blue,
citecolor=blue}

\newcommand{\ie}{\textit{i}.\textit{e}.}
\newcommand{\eg}{\textit{e}.\textit{g}.}

\def\BibTeX{{\rm B\kern-.05em{\sc i\kern-.025em b}\kern-.08em
    T\kern-.1667em\lower.7ex\hbox{E}\kern-.125emX}}

\begin{document}
\title{Federated Client-tailored Adapter for Medical Image Segmentation}
\author{Guyue Hu, Siyuan Song, Yukun Kang, Zhu Yin, Gangming Zhao, Chenglong Li, and Jin Tang
\thanks{This work was supported in part by the National Natural Science Foundation of China (No. 62376004), Natural Science Foundation of Anhui Province (No. 2408085QF201, No. 2208085J18), Natural Science Foundation of Anhui Higher Education Institution (No. 2022AH040014), The Open Project of Anhui Provincial Key Laboratory of Security Artificial Intelligence (No. SAI2024003). (Corresponding Author: Chenglong Li)}
\thanks{Guyue Hu, Siyuan Song, and Chenglong Li are with the Key Laboratory of Intelligent Computing and Signal Processing of Ministry of Education, Anhui Provincial Key Laboratory of Security Artificial Intelligence, and School of Artificial Intelligence, Anhui University, Hefei, 230601, China (e-mail: guyue.hu@ahu.edu.cn; ssy136@stu.ahu.edu.cn; lcl1314@foxmail.com)}
\thanks{Zhu Yin is with the School of Internet, Anhui University, Hefei, 230601, China (e-mail: yinzhu@ahu.edu.cn)}
\thanks{Yukun Kang and Jin Tang are with the Key Laboratory of Intelligent Computing and Signal Processing of Ministry of Education, Anhui Provincial Key Laboratory of Multimodal Cognitive Computation, and School of Computer Science and Technology, Anhui University, Hefei, 230601, China (e-mail: kyk30@stu.ahu.edu.cn; tangjin@ahu.edu.cn)}
\thanks{Gangming Zhao is with the Department of Computer Science, The University of Hong Kong, Hong Kong (e-mail:
gmzhao@connect.hku.hk)}
}

\maketitle

\begin{abstract}
Medical image segmentation in X-ray images is beneficial for computer-aided diagnosis and lesion localization. Existing methods mainly fall into a centralized learning paradigm, which is inapplicable in the practical medical scenario that only has access to distributed data islands. Federated Learning has the potential to offer a distributed solution but struggles with heavy training instability due to client-wise domain heterogeneity (including distribution diversity and class imbalance). In this paper, we propose a novel Federated Client-tailored Adapter (FCA) framework for medical image segmentation, which achieves stable and client-tailored adaptive segmentation without sharing sensitive local data. Specifically, the federated adapter stirs universal knowledge in off-the-shelf medical foundation models to stabilize the federated training process. In addition, we develop two client-tailored federated updating strategies that adaptively decompose the adapter into common and individual components, then globally and independently update the parameter groups associated with common client-invariant and individual client-specific units, respectively. They further stabilize the heterogeneous federated learning process and realize optimal client-tailored instead of sub-optimal global-compromised segmentation models. Extensive experiments on three large-scale datasets demonstrate the effectiveness and superiority of the proposed FCA framework for federated medical segmentation.

\end{abstract}


\section{Introduction}
\label{sec:introduction}

\IEEEPARstart Medical images segmentation plays a critical role in various medical applications, such as epicardial fat segmentation\cite{liu2021anatomy}, brain tumor segmentation\cite{sun2021segmentation}, facilitating more accurate diagnoses and reducing the burden on healthcare professionals. Recent progress in deep learning and large foundation models have significantly advanced the field of medical image segmentation. The pioneer U-Net \cite{unet} utilizing an encoder-decoder architecture is one of the most famous approaches in medical image segmentation. Subsequently, its variants based on various architectures have been designed to handle medical image segmentation tasks, such as CNN-based variants\cite{unet,nnunet,unet++} and transformer-based variants\cite{xie2021cotr,dalmaz2022resvit}. In recent years, numerous large medical foundation models\cite{wu2023medical,butoi2023universeg,r2} empowered with powerful capabilities of cross-domain knowledge understanding, logical reasoning, and language generation, have also significantly facilitated the field of medical image segmentation. With the aid of these approaches, medical image segmentation significantly improves the accuracy of computer-aid diagnoses and effectively streamlines clinical workflows.

\begin{figure*}[tpb]
    \centerline{\includegraphics[width=0.98\linewidth]{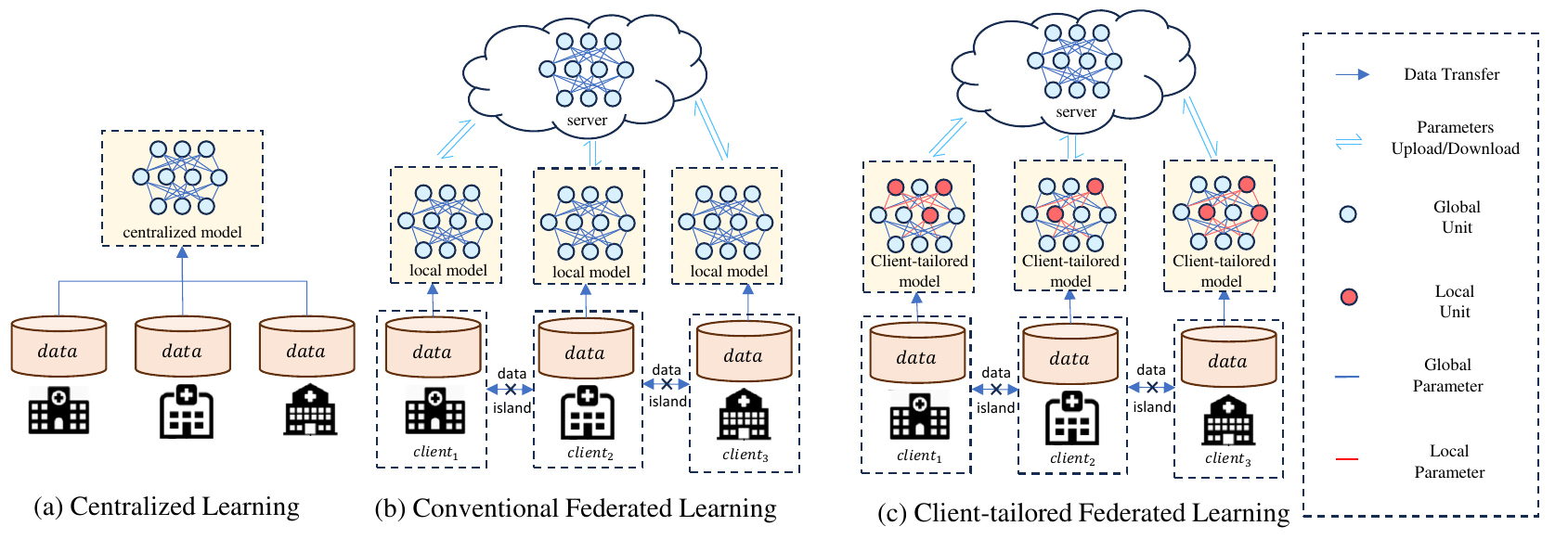}}
    \caption{Conventional learning paradigms for tackling heterogeneous distributed medical data. (a) Centralized learning aggregates all data together to train a single model. (b) Conventional federated learning trains a global-compromised model for all clients without sharing sensitive local data. (c) The proposed client-tailored federated learning trains client-customized models for each client without sharing sensitive local data.}
    \label{fig1}
    \vspace{0cm}
\end{figure*}

Despite such huge success, existing medical image segmentation methods mainly fall into a centralized learning paradigm, where medical image data from different sources (clients) are fully delivered to a central server to collectively learn a single optimal segmentation model, as shown in {Fig.}~\ref{fig1}~(a). In practical medical scenarios, we usually only have limited access to distributed ``data islands" where sharing local medical data among different clients (\eg~hospitals) is forbidden\cite{liu2019privacy} and only insensitive model weights are allowed to be shared since various factors such as strict privacy regulations in hospitals, limited network bandwidth, etc. Thus, the existing centralized approaches are no longer suitable for the distributed medical scenario. 

Federated Learning (FL) is one typical decentralized training technique, which collectively learns a global model in a central server from multiple distributed clients without sharing the distributed local data\cite{liu2021privacy,qiu2023federated,10440498}. It was initially introduced in \cite{r1}, aiming at leveraging distributed training methodologies to accommodate data from various users with disparate data scales. Since then, federated learning has achieved rapid advancements and has been applied to various fields in computer vision, such as image classification\cite{jimenez2023memory} and image segmentation\cite{jiang2023iop,miao2023fedseg}. When embracing the federated learning principle, medical image segmentation has the potential to capitalize on distributed data resources while upholding privacy regulation. As shown in {Fig.}~\ref{fig1}~(b), conventional FL first uploads local model parameters in every edge client to the global model in the central server, and then each client downloads aggregating weights from the server, eventually obtaining a global-compromised model for all clients.

\begin{figure}[tpb]
    \centerline{
    
    \includegraphics[width=1.0\columnwidth]
    {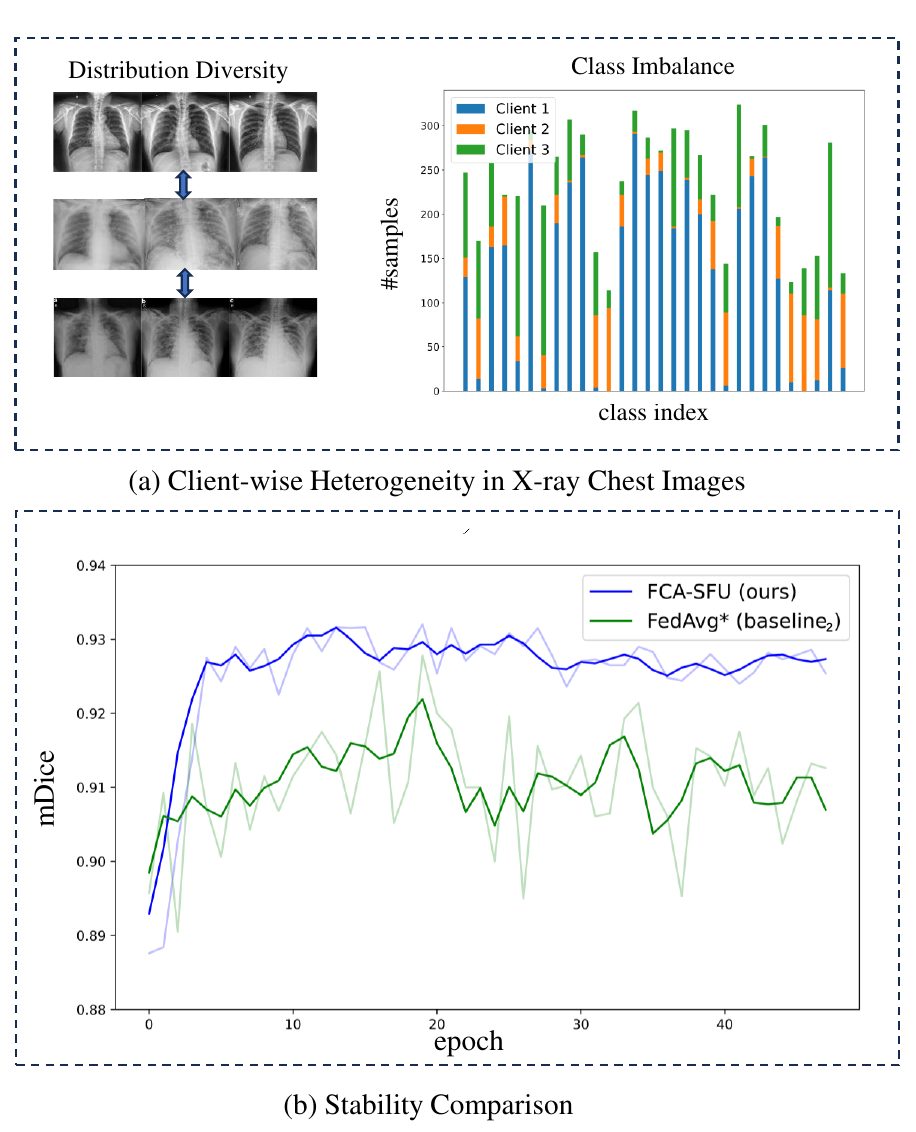}
    }
    \caption{ (a) Client-wise heterogeneity in X-ray chest images consists of common class imbalance (long-tail distribution) and various distribution diversity. (b) In heterogeneous distributed scenarios, conventional federated learning (\eg~FedAvg*\cite{r1}) suffers from considerable instability and slow convergence while our federated client-tailored adapter (FCA-SFU) effectively alleviates this issue. The transparent lines represent the original experimental results, while the solid lines represent smoothed results that facilitate visualization.
}
    \label{fig-heterogeneity}
    \vspace{0cm}
\end{figure}

However, distributed medical ``data island'' typically exhibits heavy client-wise heterogeneity including class imbalance and distribution diversity (see {Fig.}~\ref{fig-heterogeneity} (a)). Class imbalance in medical data is common for abundant reasons such as different hospitals specializing in different diseases, different anatomical regions having different probabilities being examined in medical imaging equipment, etc. The distribution diversity is usually induced by diverse data collection conditions (such as equipment, personnel, and environmental factors). Our initial experiments indicate that directly applying conventional FL methods to medical image segmentation suffers from considerable instability and slow convergence (see {Fig.}~\ref{fig-heterogeneity} (b)) since these client-wise heterogeneities. Besides, the obtained global-compromised model from conventional FL also deviates too far from the individual optimality of each client due to such heterogeneity, thus largely suppressing the advantage of utilizing distributed data for learning facilitation. Although there are a few pioneer works attempt to customize models for different clients, they usually decouple at coarse-grained levels \cite{10440498,shen2022cd2} and allocate parameter groups \cite{xie2024pflfe,chen2024towards} statically and binarily, lacking enough adaptability in tackling heterogeneous distributed medical image segmentation.

To move beyond such limitations, we propose a novel Federated Client-tailored Adapter (FCA) framework to achieve stable distributed medical segmentation without sharing sensitive local data. Specifically, we first construct parameter-efficient federated adapters to distill the client-invariant universal knowledge in off-the-shelf large medical foundation models to stabilize heterogeneous distributed medical image segmentation. In addition, We dynamically decompose the fine-grained adapter parameters into common and individual units through binary or probabilistic decomposition, as shown in {Fig.}~\ref{fig1} (c). The client-invariant components undergo a global federated updating while the client-specific individual components are updated client-independently. {The decomposed federated updating strategy} achieves two advantages: further  stabilizing the heterogeneous federated learning process and  realizing optimal client-tailored segmentation model for each client rather than sub-optimal global-compromised segmentation model for all clients.

\begin{itemize}
  \item  We identify the training instability issue in conventional federated learning induced by client-wise heterogeneity (including class imbalance and distribution diversity) in medical image segmentation and alleviate it by seeking optimal client-tailored models rather than a sub-optimal global-compromised model. 
  
  \item We propose a Federated Client-tailored Adapter for medical image segmentation, achieving stable and customized federated segmentation without sharing sensitive local data. It tightly couples parameter-efficient adapters with FL, thus effectively distilling the universal (common) knowledge in off-the-shelf medical foundation models to stabilize heterogeneous federated learning.
  \item The innovative Client-tailored Federated Updating strategies adaptively decompose the adapter units into common and individual components,  empowering a novel client-tailored and parameter-efficient updating and further stabilizing the heterogeneous federated training process. 
  \item The proposed FCA achieves state-of-the-art performance on three large-scale datasets, demonstrating its effectiveness and superiority in tackling heterogeneous distributed medical image segmentation.
\end{itemize}

\section{Related Works}
\subsection{Medical Image Segmentation}
Medical image segmentation is a crucial technique in healthcare that involves assigning each pixel in medical images with the corresponding class label. 
Existing approaches for medical image segmentation primarily fall into three paradigms: CNN-based, transformer-based, and hybridized methods. Specifically, the CNN-based methods, represented by the well-known U-Net and its variants (such as U-Net++ and nnUNet),\cite{unet,unet++,nnunet} typically employ a U-shaped encoder-decoder architecture with skip connections to preserve detailed anatomical information. These methods have demonstrated excellent performance on small-scale datasets, significantly advancing the field of medical image segmentation. Transformer-based segmentation approaches,  represented by TransUnet\cite{transunet}, leverage the robust long-range information acquisition capabilities of Vision Transformers (ViT) \cite{dosovitskiy2020image}, further improving representation capability and driving rapid advancements in medical image segmentation. Besides, VM-UNet\cite{ruan2024vm} integrates Vision-Mamba with the classical U-Net, further introducing long-distance dependencies while maintaining linear computational complexity simultaneously.

In past a few years, supervised pre-training and fine-tuning paradigm was the mainstream methods for various computer vision tasks \cite{hu2023compositional,zhu2022contrastive,hu2019joint,zhu2023information}. Beyond traditional  supervised learning paradigm, segmentation approaches that utilize foundation models and Parameter-Efficient Fine-Tuning (PEFT) techniques have also significantly reshaped and facilitated the image segmentation field. For instance, the pioneer Segment Anything Model (SAM)~\cite{r2} achieved a notable breakthrough in image segmentation by introducing a novel prompt-driven approach. The MedSAM\cite{ma2024segment} achieves high-precision segmentation across various data modalities and segmentation targets. The nnSAM\cite{li2023nnsam} combines the powerful capability of representation learning from SAM with the adaptive configuration ability from classical nnUNet \cite{nnunet}, realizing effective dataset-specific representation learning for medical image segmentation. Despite such huge success, existing segmentation methods mainly fall into a centralized learning paradigm \cite{gao2024nwpu,hu2024contrastive} and do not perform well in distributed medical scenarios.

\subsection{Federated Learning} Benefiting from the promising ability to leverage distributed data while preserving privacy, federated learning has attracted considerable attention in recent years. FedAvg \cite{r1}, one of the pioneering works in Federated Learning, offers the most fundamental framework to this paradigm. It employs a simple weighted averaging strategy to update a global model in the central server. However, recent studies have highlighted the phenomenon of client drift induced by client-wise heterogeneity \cite{gao2022feddc,miao2023fedseg}, resulting in inconsistency issues regarding optimal models for each client. Therefore, some improvements have been developed to alleviate the client-wise non-iid challenge. The FedProx \cite{FedProx}, FedDC \cite{gao2022feddc}, FedSM \cite{xu2022closing}, and FedSeg \cite{miao2023fedseg} reform the federated aggregation strategy or loss functions to seek more matchable global or local models. FedNH \cite{dai2023tackling} and Fed-CBS \cite{zhang2023fed} alleviate client-wise class imbalance by a client sampling strategy to generate grouped class-balanced datasets and utilizing the uniformity and semantics of class prototypes, receptively. Scaffold \cite{karimireddy2020scaffold}, and FedNP \cite{wu2023fednp} alleviate non-iid data issues by deliberately handling distribution diversity in distributed datasets. FedST \cite{ma2024fedst} and FedA3I \cite{wu2024feda3i} enhance the contribution of high-quality local models during aggregation, aiming to achieve superior aggregated models. FSAR \cite{guo2023fsar} divides the topology connection in the graph neural network into global and local parts, realizing adaptive federated action recognition. FedNH \cite{dai2023tackling} improves the client drift phenomenon during aggregation by enhancing the generalization of local models. In our work, we develop a novel client-tailored federated updating strategy that adaptive and fine-grained decomposes global client-invariant and local client-specific units, realizing optimal client-customized models for each client rather than a sub-optimal global-compromised model for all clients.

\begin{figure*}[htbp]
    \centerline{\includegraphics[width=0.94\linewidth]{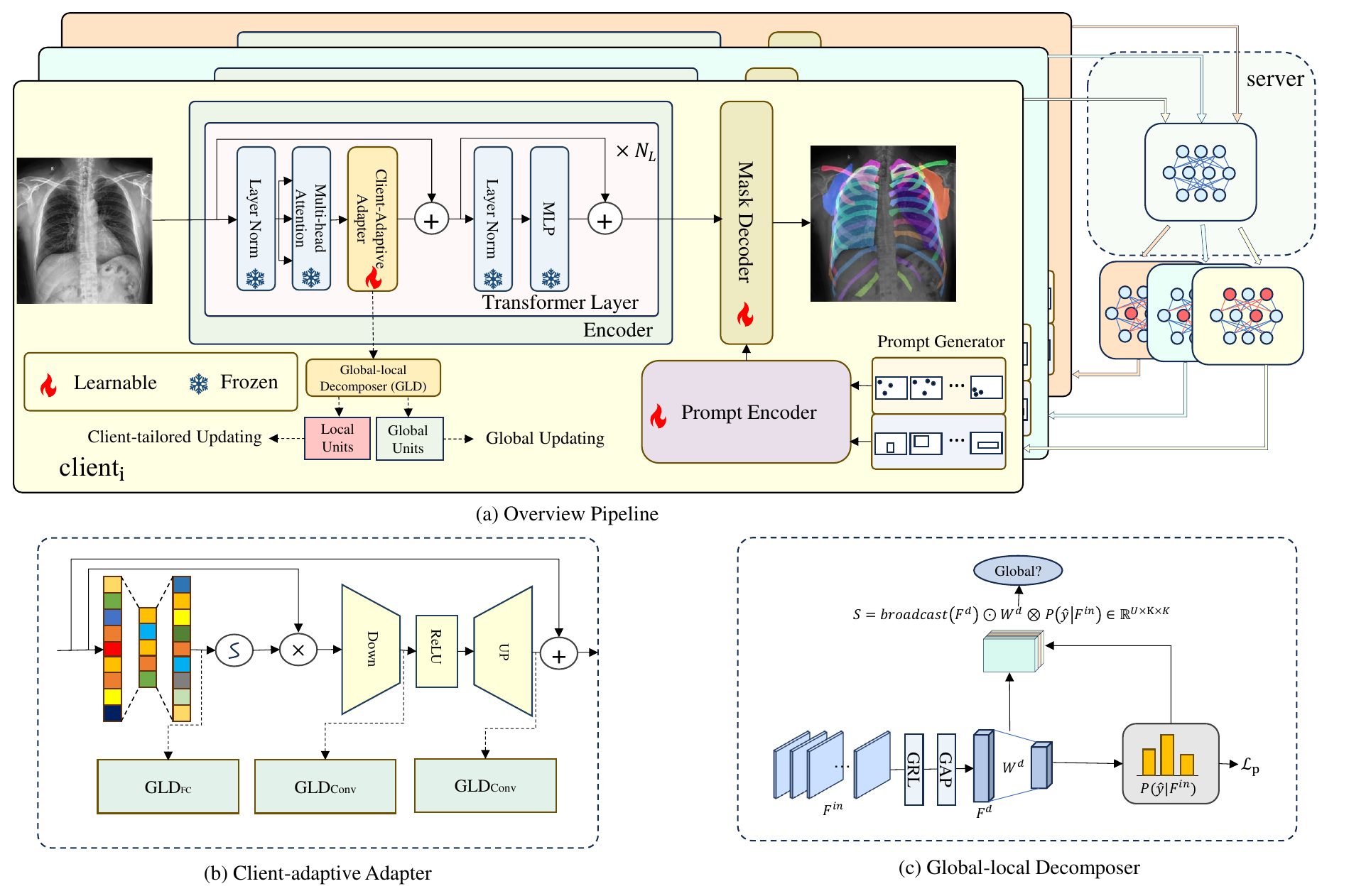}}
    \caption{ (a) Overview of the proposed federated client-tailored adapter (FCA) framework. (b) Detailed structure of the client-adaptive adapter. (c) The mechanism of the Global-local decomposer (GLD).
    }
    \label{Overview}
    \vspace{-0cm}
\end{figure*}

\section{Method}
\subsection{Preliminary}
    \subsubsection{Conventional Federated Learning} 
    We first introduce a vanilla baseline under the umbrella of conventional federated learning (FL) to demonstrate some preliminary knowledge, illustrated in {Fig.}~\ref{fig1} (b). Consider a distributed medical scenario with \(N\) edge clients and one central server, where each client holds some private data that cannot be shared among clients and the central server. During training, all clients collaboratively update a shared segmentation model in the central server. For each global updating round, every client trains its latest local model received from the server for \(N_{e}\) epochs based on its private data. Then, each local client sends the updated parameters to the central server for aggregating and updating the global segmentation model. The updated global model is subsequently distributed to each local client for parameter replacement and waiting for the next updating round. As a result, conventional FL usually leads to a \textit{global-compromised} segmentation model for \textit{all clients} rather than \textit{client-tailored} segmentation models for \textit{each client}, and each client utilizes the same model during the reference period. Moreover, conventional FL also encounters considerable instability and slow convergence issues induced by client-wise heterogeneity (class imbalance and distribution diversity) in distributed medical data (see {Fig.}~\ref{fig-heterogeneity}~(a) for details). In this paper, we proposed a Federated Client-tailored Adapter (FCA) for medical image segmentation to address these issues.
    
    \subsubsection{Medical Foundation Models} In the last few years, large medical foundation models (MFMs) have significantly advanced various tasks in the medical image processing field. For example, one of the most famous MFMs is SAM-Med2D~\cite{r2}, which fine-tunes conventional segmentation model SAM~\cite{kirillov2023segment} via Parameter-Efficient Fine-Tuning (PEFT) techniques of adapter learning and prompt learning \cite{lialin2023scaling}. Most MFMs for medical image segmentation (such as  MedSAM\cite{ma2024segment} and Med-SA\cite{zhang2024segment}) usually follow a similar pipeline that includes a transformer-based image encoder, a transformer-based mask decoder, a prompt generator, and a prompt encoder. The encoder layers are implemented with a Vision Transformer (ViT) that extracts image features through multiple stacked transformer layers. A prompt encoder encodes the information of prompt hints generated from text, points, or boxes. Finally, the mask decoder integrates the representations from the image encoder and the prompt encoder to generate corresponding segmentation masks.  For simplicity, we take the famous SAM-Med2D~\cite{r2} as an example to demonstrate our FCA framework in the method section and validate its generalization ability in the experimental section (Section \ref{sec:exp}). The off-the-shelf MFMs inherently contain abundant general medical knowledge because of their huge model parameters and massive training data. Therefore, it is potential to distill the client-invariant universal knowledge from off-the-shelf MFMs as one effective measure to stabilize the non-centralized medical image segmentation.

\subsection{Overview}
The overview pipeline of our Federated Client-tailored Adapter (FCA) for tackling heterogeneous medical image segmentation is shown in {Fig.}~\ref{Overview}~(a). The basic segmentation model in each local client consists of a large medical foundation model (specified as the SAM-Med2D in {Fig.}~\ref{Overview}~(a)) and inserted adapter layers in each transformer layer of the MFMs encoder. Most layers of the MFMs encoder are frozen, which contain rich prior knowledge (including both client-specific and client-invariant knowledge) inherited from off-the-shelf MFMs. Only the lightweight adapter layers, prompt encoder, and mask decoder with a few learnable parameters will go through parameter-efficient fine-tuning. As a result, each local client can efficiently distill client-specific dark knowledge in the MFMs to facilitate its client-tailored knowledge learning process. They also stir client-invariant general knowledge to benefit and stabilize the sequential client-invariant federated updating process.

Then, the central server and all edge clients fed with the above basic segmentation models conduct federated learning with heterogeneous distributed medical ``data islands". If directly applying conventional federated updating strategy ( {Fig.}~\ref{fig1}~(b)), it could only obtain a \textit{global-compromised} segmentation model and also encounters considerable instability and slow convergence. Therefore, we develop a Global-local decompose mechanism (GLD) to adaptively decompose each adaptor into client-invariant global units and client-specific local units. Thereafter,  the proposed client-tailored federated updating strategies are explored to binary or smoothly to update the adapters in the local clients and central server. As a result, our FCA framework could obtain more optimal \textit{client-tailored} segmentation models for \textit{each client} rather than a sub-optimal \textit{global-compromised} segmentation model for \textit{all clients}. Moreover, our client-tailored federated updating strategies also alleviate training instability and slow convergence issues during heterogeneous federated learning.

\begin{figure}[!t]
\centerline{\includegraphics[width=\columnwidth]{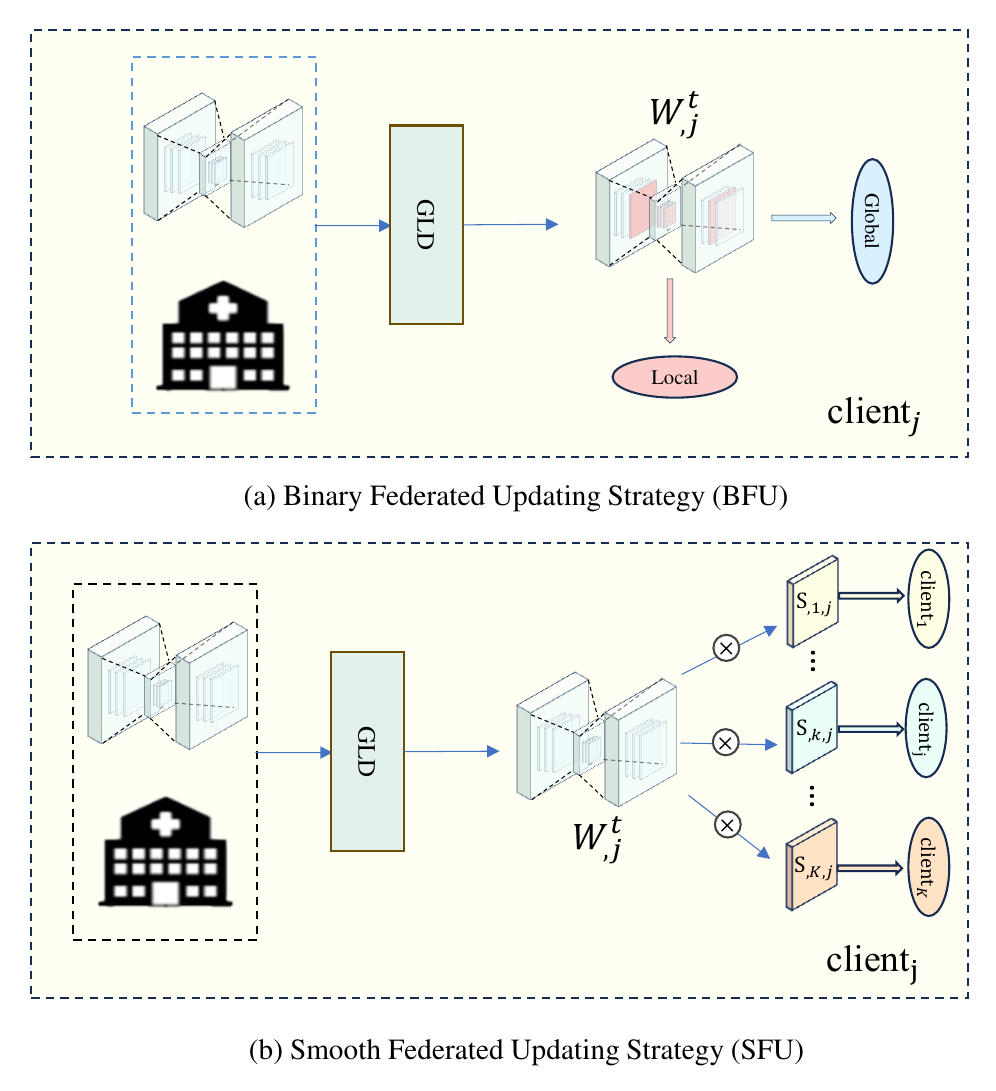}}

\caption{(a) The binary federated updating (BFU) strategy binary distinguishes the adapter units in each client into local client-specific and global client-invariant units and respectively updates them locally and globally. (b) The smooth federated updating (SFU) strategy probabilistically distributes each unit to all clients and thus each unit probabilistically participates in the federated parameter updating of all clients.}

\label{fig4}
\vspace{0cm}
\end{figure}

\subsection{Adaptive Adaptor Decomposition}

As shown in {Fig.}~\ref{Overview}~(b), a typical adapter (borrowed from \cite{r2}) for large MFMs usually consists of a series of fully connected (FC) and convolutional (Conv) layers. We append an auxiliary Global-local Decomposer (GLD) branch to each Conv and FC layer to adaptively decompose the adaptor into global and local units, which are denoted as \(GLD_{Conv}\) and \(GLD_{FC}\) in {Fig.}~\ref{Overview}~(b). The Global-local Decomposer aims to decompose the units (channel for Conv layers or neuron for FC layers) in each adapter layer into common client-invariant and individual client-specific components. The parameters group associated with global and local units will be updated globally and locally thereafter. Taking the \(GLD_{Conv}\) as an example, we describe its implementation details in {Fig.}~\ref{Overview}~(c). The \(GLD_{Conv}\) comprises a lightweight client discriminator that distinguishes which client (domain) the input representation comes from. Specifically, the side input representation \(\bm{F}^{in}(i) \in \mathbb{R}^{ H \times W \times C}\) corresponding sample \(i\)  first goes through a global average pooling (GAP) layer along the spatial dimensions to obtain a channel-wised presentation \(\bm{F}^{d}(i) \in \mathbb{R}^{C}\), where \(C\) is the channel number of the feature map. 
Then, we construct a pretext task called client discrimination to dynamically distinguish the representation source and determine the individual unit contribution during each global communication round. The client discriminator takes in representation from different clients and will be trained to distinguish the representation source. Given an input representation \(\bm{F}^{in}(i)\) of sample \(i\)  from one specific client, the client discriminator yields a classification probability \(P(\hat{\bm{y}}_k(i)|\bm{F}^{in}(i))\) voting it sources from the \textit{k}-th client. The train loss for this pretext task is defined as follows, 
\begin{equation}
\label{eq1}
\mathcal{L}_p = \frac{1}{N_I} \sum_{i=1}^{N_I}\sum_{j=1}^K \mathbbm{1}_{[j=k]} \log{(\bm{W}^d (GAP(\bm{F}^{in}_{k}(i))))}
\end{equation}
where $N_I$ and $K$ are respectively the numbers of samples and clients, and \(\mathbbm{1}\) denotes the indicator function. To ensure the auxiliary branch does not affect the original training process of the main branch in {Fig.}~\ref{Overview}~(b), we elaborately insert a Gradient Reversal Layer (GRL) before the client discriminator to truncate the gradients from the above discrimination loss.

Taking one convolutional layer in the MFMs adapter as an example, we treat each channel as a unit and let $U=C$. Intuitively, the channels contributing more to client prediction contain more client-specific information while those contributing less may contain more client-invariant global information. Thus, we first quantify the contribution score $\bm{\hat{S}}_{u,k}(i)$ of each channel (unit) \(u\) utilized for determining an input feature $\bm{F}^{in}(i)$ corresponding to sample \(i\) is sourced from client \(k\) via its weighted activation and then weight it by predicted client probability \(P(\hat{\bm{y}}_j^i|\bm{F}^{in}(i))\). The whole process of computing weighted the contribution score could be formulated as \(\bm{S}_{u,k,j}(i)\) = \(\bm{F}_u^{d}(i) \cdot \bm{W}_{u,k}^d \cdot P(\hat{\bm{y}}_j^i|\bm{F}^{in}(i))\). For brevity, we rewrite the process in matrix form, \ie
\vspace{0cm}
\begin{equation}
\label{eq2}
\hat{\bm{S}}(i) = broadcast(\bm{F}^d(i)) \odot \bm {W^d}
\vspace{0cm}
\end{equation}

where $\hat{\bm{S}}(i) \in \mathbb{R}^{U\times{K}}$, the function \(broadcast\) denotes element duplicating for shape matching, \(\bm{W}^d \in \mathbb{R}^{U\times{K}}\) is the classifier weight of the GLD (its various elements can represent the contribution level of each unit judged as a certain category), and \(\odot\) is element-wised multiplication. \(\hat{\bm{S}}(i)\) represents the contribution of each unit to determining whether the entire feature belongs to a certain category. After weighing the previous contribution score of each unit with the predicted domain probability \(P(\hat{\bm{y}}^i|\bm{F}^{in}(i)) \in \mathbb{R}^{{K}} \) corresponding to the input feature, we obtain the weighted contribution score matrix \(\bm{S}(i)\) corresponding sample \(i\),
\vspace{0cm}
\begin{equation}
\label{eq3}
\bm{S}(i) = \hat{\bm{S}}(i) \otimes P(\hat{\bm{y}}^i|\bm{F}^{in}(i))
\vspace{0cm}
\end{equation}
where the weighted contribution score \(\bm{S} (i) \in\mathbb{R}^{U\times{K}\times{K}}\), and  $\otimes$ represents the outer product. Finally, we conduct a sample-wise average to obtain the final matrix of contribution score, \ie~$\bm{S}$ = \(\sum_{i=1}^{N_I}\bm{S}(i)\), where $N_I$ is the number of samples. To maintain brevity, we will exclude the sample index $i$ for all related matrices from here unless explicitly stated.

As for the implementation of a Global-local Decomposer for FC layers (\(GLD_{FC}\)), we could treat it as a special convolution layer with a kernel of \(1\times{1}\). Then, the \(GLD_{FC}\) could be implemented similar as the \(GLD_{Conv}\). Eventually, the final contribution score $\bm{S}$ will be delivered to the central server and then broadcast to every client for global and local unit discrimination and client-tailored federated updating.

\subsection{Client-tailored Federated Updating}
\label{section-update}
 
 In this section, we split the adapter units into common client-invariant and individual client-specific components according to the weighted contribution score \(\bm{S}\). Then, we conduct global updating for common client-invariant units and independent updating for individual client-specific units. Specifically, we first establish a model copy in the central server including all components of each local client in {Fig.}~\ref{Overview}~(a) except for the insertion of the client-tailored adapters. Then, we devise two client-tailored federated updating strategies to conduct client-tailored federated learning, including binary federated updating (BFU) and smooth federated updating (SFU). The SFU is conceptually built upon BFU and could be treated as a generalized version of BFU. Our experiments indicate that both strategies are very effective in alleviating training instability and slow convergence issues during heterogeneous federated learning, among which the generalized strategy SFU achieves better performance. The detailed algorithm is shown in {Algorithm}~\ref{alg:cap}. We will introduce the implementation details below.

\begin{algorithm}
\caption{Client-tailored Federated Updating (FCA)}\label{alg:cap}
\begin{algorithmic}
\State \textbf{Input:} The number of input samples $N_I$, the number of clients $K$, the learnable parameters of encoders in each client $\bm{W}$, the unit number in a adapter layer \(U\)

\State \textbf{Output:} Client-tailored adaptor for every client ${W}^{T+1}(\theta,\hat{\theta}$)

\For{$t=1$ to $T$}
    \\\textit{// Update local models and get contribution score \(\bm{S}\)}
    \For{$k=1$ to $K$}
        \State $\bm{W}_k^{t+1}(\theta,\hat{\theta})  \gets \bm{W}_k^t(\theta,\hat{\theta}) - \eta \bigtriangledown l (\bm{W}_k^t)(\theta,\hat{\theta}) $ 
        \State \textit{// \(\theta\), \(\hat{\theta}\) denotes global and local updating parameters}
        
            \State \(\bm{S}(i) \gets {broadcast(\bm{F}^d(i)) \odot \bm {W^d} \otimes P(\hat{\bm{y}}^i|\bm{F}^{in}(i))}\)
            
            \State \textit{//{Eq.}~\ref{eq2}~\&~{Eq.}~\ref{eq3}}
            \State \(\bm{S} \gets \frac{1}{N_I}\sum_{i=1}^{N_I}S(i)\)
        
    \EndFor
    \State\textit{// Binary Federated Updating Strategy (BFU)}
    \If {\text{apply}\  \text{BFU Strategy}}
    
        \For{$u,k=(1,1)$ to $(U,K)$}
        \State 
            \State $\bm{D}_{u,k} \gets \dfrac{\sum_{j=1}^K \bm{S}_{u,k,j}\times\log_{2}{\bm{S}_{u,k,j}}}{\log_{2}{K}}\ $  \textit{// {Eq.}~\ref{eq4}}
            
            \State $\bm{M}_{u,k} \gets 
            \begin{cases}
                \text{1 } \ \ \ \bm{D}_{u,k} > \delta \\
                \text{0 } \ \ \ \bm{D}_{u,k} < \delta
            \end{cases}$\textit{// {Eq.}~\ref{eq5}}
            \State 
        \EndFor
        \State \textit{// Integration$(\bm{W}_1^t(\theta),\bm{W}_2^t(\theta),\cdots,\bm{W}_k^t(\theta))$}
        \For{$u,k=(1,1)$ to $(U,K)$}
            \If{$\bm{M}_{u,k} == 1$}
            
                \State $ \bm{W}_u^{t+1} \gets \dfrac{\sum_{k=1}^K \bm{M}_{u,k}\times{{\bm{W}_{u,k}^t}}}{\sum_{k=1}^K \bm{M}_{u,k}}\ $\textit{// {Eq.}~\ref{eq6}}
                 
            \EndIf
                
        \EndFor

    \EndIf
    \State\textit{// Smooth Federated Updating Strategy (SFU)}
    \If{\text{apply}\  SFU Strategy}

        \State\textit{// Integration$(\bm{W}_1^t(\theta,\hat{\theta}),\bm{W}_2^t(\theta,\hat{\theta}), \cdots, \bm{W}_k^t(\theta,\hat{\theta})) $}
        \For{$u,j,=(1,1)$ to $U,K$}
                
            \State $ \bm{W}_{u,j}^{t+1} = \dfrac{\sum_{k=1}^K \bm{S}_{u,k,j}\times{\bm{W}_{u,k}^t}}{\sum_{k=1}^K \bm{S}_{u,k,j}}\ $\textit{// {Eq.}~\ref{eq7}}

        \EndFor
    \EndIf
\EndFor
\end{algorithmic}
\end{algorithm}

\begin{table*}[]
\begin{center}
\caption{Comparing with the state-of-the-art and baseline methods regarding mDice on the CXRS-HG and HLS datasets}
\begin{tabular}{l|c|ccc|c|ccc}
\hline \hline
\multirow{2}{*}{\textbf{Methods}} & \multicolumn{4}{c|}{\textbf{CXRS-HG}}                                         & \multicolumn{4}{c|}{\textbf{HLS}}                                             \\ \cline{2-9} 
                                 &
                                 \textbf{Average} &
                                 \textbf{Client\textsubscript{1}} & \textbf{Client\textsubscript{2}} & \textbf{Client\textsubscript{3}} &
                                 \textbf{Average}&
                                 \textbf{Client\textsubscript{1}} & \textbf{Client\textsubscript{2}} & \textbf{Client\textsubscript{3}} 
                                 
                                 \\ \hline 
FedAvg\cite{r1} (baseline\(_1\))                  & 36.94         & 35.82            & 37.71            & 37.29            & 58.41            & 58.39            & 58.80            & 58.04                        \\
FedProx\cite{FedProx}                     & 37.74     & 37.40            & 37.97            & 37.85            & 57.20            & 56.81            & 57.47            & 57.32                        \\
HarmoFL\cite{jiang2022harmofl}                 & 42.85         & 42.73            & 42.97            & 42.85            & 51.13            & 50.27            & 51.52            & 51.60                        \\
FedSeg\cite{miao2023fedseg}             & 40.35               & 38.42            & 41.56            & 41.07           & 81.72            & 79.13            & 82.69            & 83.34                       \\ 
FedA3I\cite{wu2024feda3i}            &41.99              & 53.75            & 34.02            & 38.22            & 81.58            &  80.98           & 83.30            & 80.48                       \\ \hline
FedAvg*\cite{r1} (baseline\(_2\))            & 60.64             & 60.58            & 60.69            & 60.67            & 90.79            & 88.76            & 90.87            & 92.75                        \\
FedProx*\cite{FedProx}             & 61.06            & 61.27            & 61.48            & 60.43            & 91.06            & 88.80            & 91.38            & 92.84                        \\

PerFedAvg*\cite{zhou2024personalized}             & 61.35             & 61.11            & 61.39                        & 61.56       & 91.37     & 89.95            & 91.28           & 92.90                        \\
FedCross*\cite{xu2024federated}             & {61.44}             & {61.09}            & {61.75}                        & {61.29}       & {91.03}     & {89.21}            & {91.01}           & {92.87}                        \\
MAP*\cite{li2024map}             & 61.54             & 61.22            & 61.39                        & 62.01       & 91.49     & 90.18            & 91.54           & 92.75                       \\

IOP-FL*\cite{jiang2023iop}             & 62.00             & 61.21            & 61.86                        & 61.69       & 91.68     & 90.34            & 91.96           & 92.69                        \\ \hline
FCA-BFU (\textbf{ours})      & \textbf{63.41}     & \textbf{63.31}            & \textbf{63.20}            & \textbf{63.15}            & \textbf{92.12}            & \textbf{90.60}            & \textbf{92.26}            & \textbf{93.50}                       \\
FCA-SFU (\textbf{ours}) &\textbf{64.15} & \textbf{64.08}   & \textbf{64.47}   & \textbf{63.90}    & \textbf{92.44}   & \textbf{90.82}   & \textbf{92.52}   & \textbf{93.98}      \\ \hline \hline
\end{tabular}

\vspace{0.1cm}

\item[*] Our implementation of MFMs enhanced variants of conventional FL methods.

\label{table1}
\end{center}
\vspace{0cm}
\end{table*}

\subsubsection{Binary Federated Updating Strategy}
In each round, the Binary Federated Updating Strategy (BFU) binary distinguishes the adapter units in each local model into client-specific local parts and client-invariant global parts ({Fig.}~\ref{fig4}~(a)). Then, the parameter group corresponding to client-invariant global parts will communicate with the central server for joint global updating while the parameter group corresponding to client-specific local parts only be updated locally. The weighted contribution score \(\bm{S}\) contains the contribution of each unit for client discrimination and serves as a good indicator to determine whether the units should be treated locally or globally. Intuitively, the adapter units that contribute more to client prediction likely contain more client-specific information. In contrast, the units that contribute less contain more client-invariant global information. To determine whether the parameter group in a unit should be treated globally or locally, we examine the uniformity of its weighted contribution scores for client discrimination. If the weighted contribution scores are similar across all clients, it suggests that the parameter group in this unit is client-specific and thus can be considered a global parameter. In contrast, if there are significant score differences across clients, it indicates that the parameter group in the unit may be domain-specific and should be treated locally. Therefore, we quantify the client-wise uniformity of the weighted contribution score by computing the normalized entropy of score distribution, \ie  
\vspace{0cm}
\begin{equation}
\label{eq4}
\bm{D}_{u,k} = \dfrac{Entropy(\bm{S}_{u,k})}{Entropy(\mathcal{U}(1, K))}=\dfrac{\sum_{j=1}^K \bm{S}_{u,k,j}\times\log_{2}{\bm{S}_{u,k,j}}}{\log_{2}{K}}\ 
\vspace{0cm}
\end{equation}

where \(\bm{D}_{u,k}\) represents the diversity degree of score distribution for the unit \textit{u} of the \textit{k}-th client, and the \(\mathcal{U}\) represents the uniform distribution. The \(\log_{2}{K}\) is the theoretical upper bound for the entropy of client-wise score distribution when the distribution is a standard uniform distribution (\ie~\(Entropy(\mathcal{U}(1, K))=\log_{2}{K}\)). As a result, the larger the diversity degree of a unit, its distribution is closer to a uniform distribution, and its parameter group is more likely to be global. Therefore, we could conveniently distinguish whether each adapter unit is global (client-invariant) or local (client-specific) through binarizing $\bm{D}_{u,k}$ with a threshold \(\delta\) (see  {Fig.}~\ref{fig4}~(a) for details). If \(\bm{D}_{u,k}\) for a unit $u$ of client $k$ is greater than \(\delta\), the unit is treated as a global unit. Otherwise, if \(\bm{D}_{u,k}\) is less than \(\delta\), the unit is treated as a local unit. The detailed formula is as follows:
\vspace{0cm}
\begin{equation}
\bm{M}_{u,k} = 
\begin{cases}
\label{eq5}
   1,  \quad \bm{D}_{u,k} > \delta \quad & (Global\ Unit)\\
   0,  \quad \bm{D}_{u,k} < \delta \quad & (Local\ Unit) 
\end{cases}
\vspace{0cm}
\end{equation}
where the mask \(\bm{M}_{u,k}\) denotes whether an adapter unit $u$ of the \textit{k}-th client is global or not.

As shown in {Algorithm}~\ref{alg:cap}, our binary federated updating (BFU) strategy applies different updating processes for the local and global units during federated updating. As for global units (\(\bm{M}_{u,k}=1\)) in a particular client, their parameter group will be updated as conventional FL. Specifically, the parameters in a local client will be uploaded to the central server and averaged with corresponding parameters from other clients. Then, the average value will be sent back to the local client for updating. The updating process for global units could be formulated as follows:
\vspace{0cm}
\begin{equation}
\label{eq6}
\bm{W}^{t+1}_u = \dfrac{\sum_k^K \bm{M}_{u,k}\times{{\bm{W}_{u,k}^t}}}{\sum_k^K \bm{M}_{u,k}}\
\vspace{0cm}
\end{equation}
where \(t\) represents the current updating round, and \(t+1\) represents next round. In contrast,  if the unit is identified as a local unit (\(\bm{M}_{u,k}=0\)), its parameter group is only updated locally and does not undergo any global updating process. Eventually, our BFS advances the conventional federated learning approaches by enabling binary parameter categorization (\ie~client-invariant global parameters and client-specific local parameters), then realizing client-tailored federated learning for heterogeneous medical image segmentation. The detailed updating process of the proposed binary federated updating (BFU) strategy is shown in {Algorithm}~\ref{alg:cap}.

\begin{table}[]
\begin{center}
\caption{Comparing with the state-of-the-art and baseline methods regarding mDice on the AMD-SD-HG dataset}
\begin{tabular}{l|c|ccc}
\hline \hline
\textbf{Methods}
                                 &
                                 \textbf{Average}&
                                 \textbf{Client\textsubscript{1}} & \textbf{Client\textsubscript{2}} & \textbf{Client\textsubscript{3}}
                                 
                                 \\ \hline 
FedAvg\cite{r1} (baseline\(_1\))      &45.12&44.25&46.03&45.1                  \\
 \hline
FedAvg*\cite{r1} (baseline\(_2\)) &60.42&59.55&61.42&60.31                    \\
FedCross*\cite{xu2024federated}&60.71&60.56&61.01&60.56               \\
MAP*\cite{li2024map} &61.1&60.96&61.65&60.69             \\

IOP-FL*\cite{jiang2023iop}&61.41&61.24&61.88&61.13      \\ \hline
FCA-BFU (\textbf{ours})&\textbf{62.56}&\textbf{62.15}&\textbf{63.12}&\textbf{62.41}        \\
FCA-SFU (\textbf{ours}) &\textbf{63.40}&\textbf{62.5}&\textbf{64.47}&\textbf{63.24}   \\ \hline \hline
\end{tabular}

\vspace{0.1cm}

\item[*] Our implementation of MFMs enhanced variants.

\label{compare_2}
\end{center}
\vspace{0cm}
\end{table}

\begin{figure*}[!t]

\centerline{\includegraphics[width=0.93\textwidth]{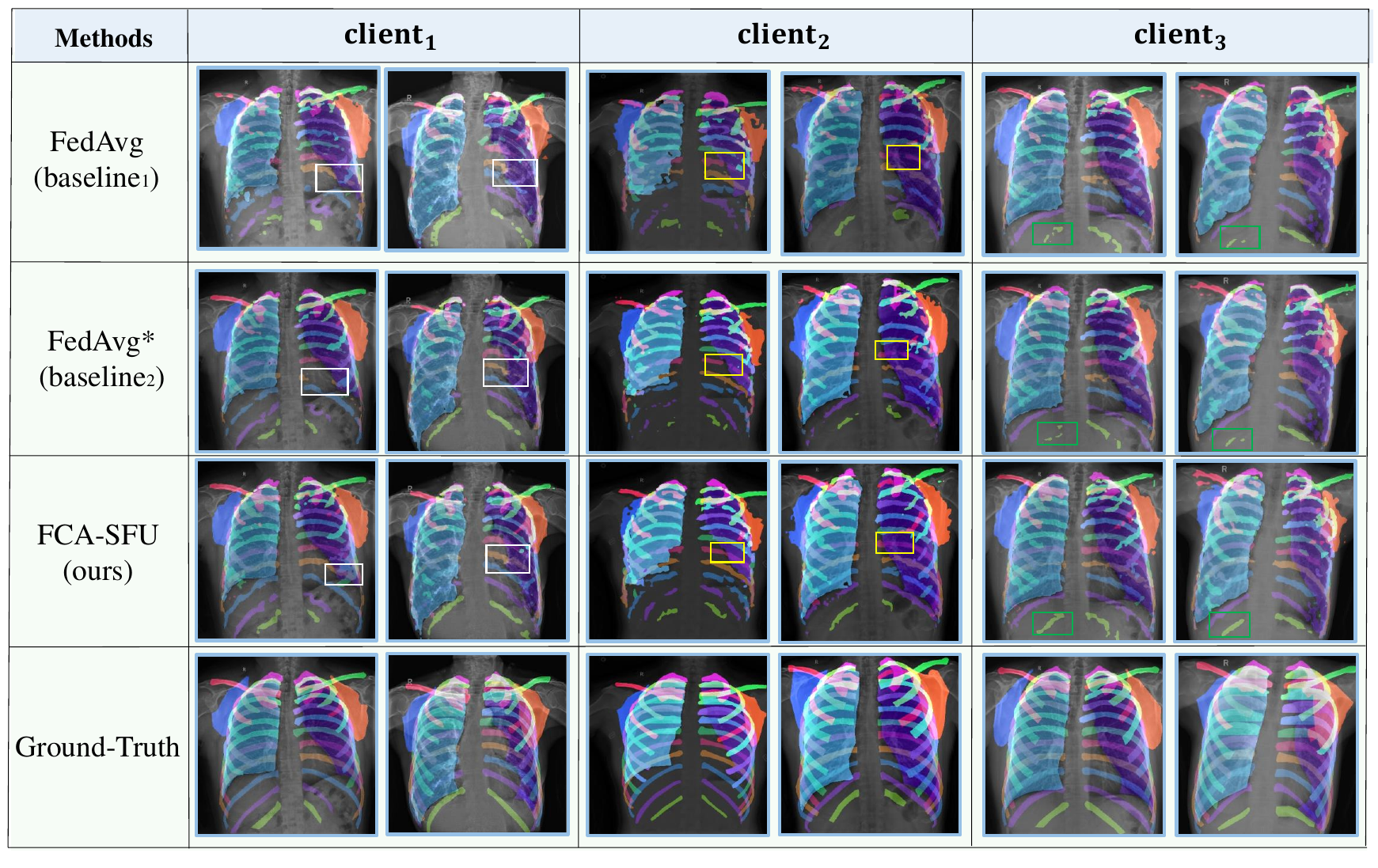}}

\caption{Visualization comparison of heterogeneous federated segmentation results on the CXRS-HG dataset}

\label{visualization}
\vspace{0cm}
\end{figure*}

\subsubsection{Smooth Federated Updating Strategy}
Although the above BFU strategy already realizes client-tailored federated learning through binary parameter categorization, it is somewhat rigid and not good enough for complex scenarios where parameters simultaneously encapsulate complicated information from multiple client domains. To solve such limitation, we further proposed a Smooth Federated Updating (SFU) strategy. It is conceptually built upon the above BFU and could be treated as a more generalized version of BFU because it utilizes a smoother parameter association substituting the binary parameter categorization. Specifically, unlike the binary strategy that assigns each unit exclusively to be local or global, our smooth federated updating strategy considers each unit probabilistically (rather than exclusively) belonging to a client. As shown in {Fig.}~\ref{fig4}~(b), we exploit the weighted contribution score of each unit on different clients to weight the parameter group of each unit on each client, generating a unit-wise weighted model for each client. Then, each client exploits the weighted models of all clients to update itself. Specifically, when updating the parameter group \(\bm{W}_{u,j}^t\) of unit \(u\) on client \(j\) at updating round $t$, we use the probabilistic contribution score for every client to weight the parameter updating from every client \(k\), \ie
\vspace{0cm}
\begin{equation}
\label{eq7}
\bm{W}_{u,j}^{t+1} = \dfrac{\sum_{k}^K \bm{S}_{u,k,j}\times{\bm{W}_{u,k}^t}}{\sum_{k}^K \bm{S}_{u,k,j}}.
\vspace{0cm}
\end{equation}

As a result, our SFU incorporates the weighted contribution score regarding every client as smoothly adjusting weights for federated aggregating, which  enables a more adaptive and dynamic updating of model parameters and realizes smooth client-tailored segmentation for each client. It allows the parameter update in a local client contributing to the global aggregating in proportion to its relevance to other clients. 

In summary, our BFU strategy offers a straightforward parameter categorization pipeline for heterogeneous federated learning. The SFU further leverages probabilistic weighting based on client-specific scores, providing a more nuanced and flexible mechanism for parameter management during federated updating, significantly enhancing the performance of heterogeneous federated segmentation.

\begin{figure*}[!t]
\centerline{\includegraphics[width=0.93\textwidth]{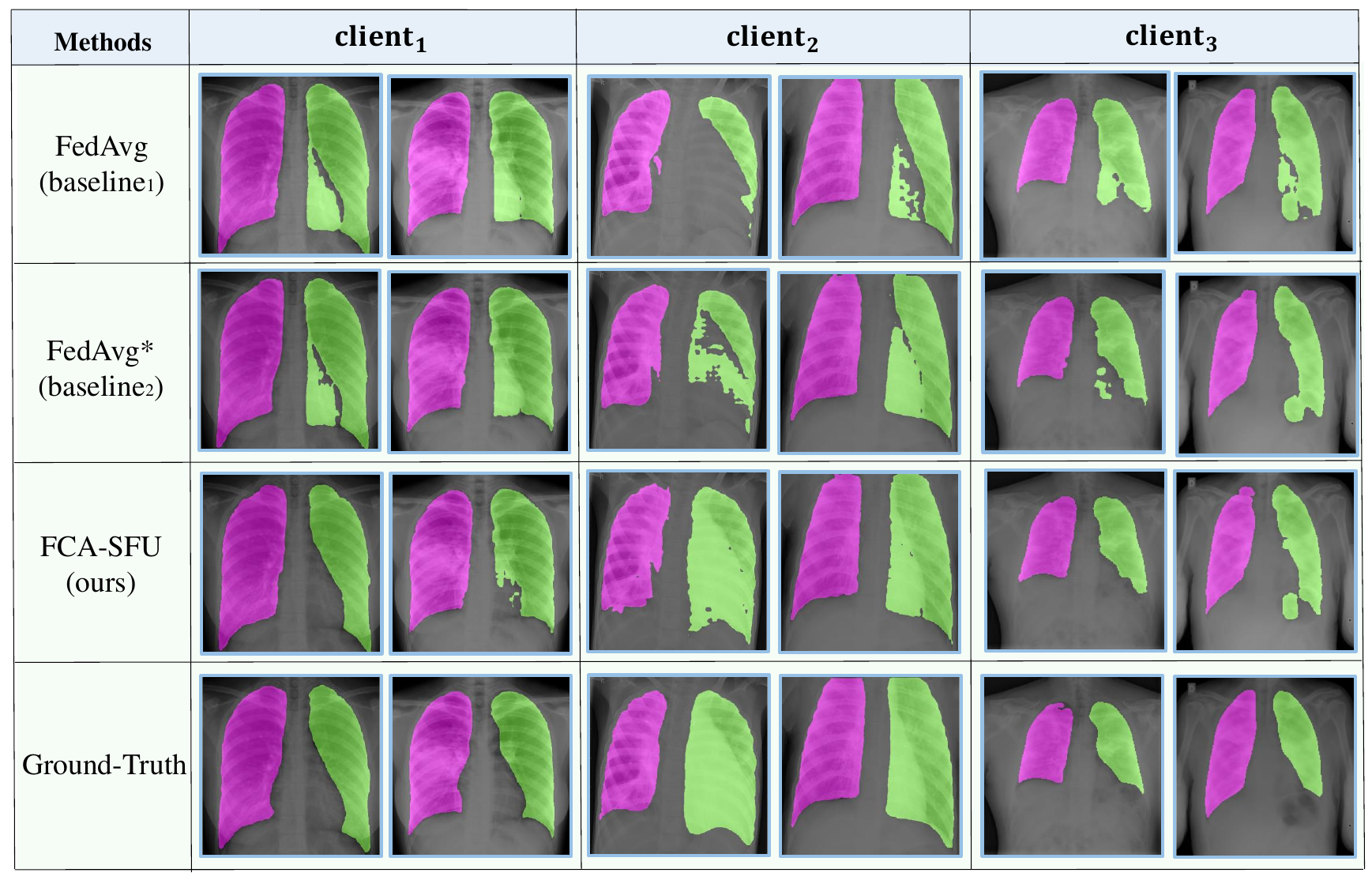}}

\caption{Visualization comparison of heterogeneous federated segmentation results on the HLS dataset}

\label{visualization2}
\vspace{-0.0cm}
\end{figure*}

\section{EXPERIMENTS} 
\label{sec:exp}
\subsection{Dataset}
\subsubsection{CXRS-HG dataset}The CXRS-HG dataset is a heterogeneous distributed dataset constructed from the Chest X-ray Segmentation (CXRS) dataset \cite{10884681}. Original CXRS is an in-house dataset comprising 1250 chest X-ray images of 30 different anatomical structures (including 24 ribs, 2 clavicles, 2 scapulae, and 2 lungs) for medical image segmentation, and each image is annotated with 30 anatomical segmentation masks. We reform the centralized CXRS dataset into a distributed dataset CXRS-HG containing client-wise heterogeneity. Following the classical protocol for federated learning in \cite{mu2023fedproc}, we first distribute the image samples to distributed clients following a standard Dirichlet distribution to simulate the heterogeneity of class imbalance in a practical medical scenario (as shown in {Fig.}~\ref{fig-heterogeneity}~(a). Then, the distributed X-ray images in different local clients will undergo different image transformation strategies (original images for client\textsubscript{1}, 3x3 mean blur filtering for client\textsubscript{2}, half down-sampling and then restoring resolution for client\textsubscript{3}) to simulate the heterogeneity induced by various environmental factors (such as surroundings and imaging devices). As a result, the distributed CXRS-HG contains abundant client-wise heterogeneity including class imbalance and distribution diversity which is highly similar to non-centralized medical segmentation scenarios. Following \cite{ma2024fedst,wang2023feddp}, we employ the mean Dice Similarity Coefficient (mDice) as the metric to evaluate segmentation performance on the CXRS-HG dataset, which is the class-wise mean of the Dice Similarity Coefficient for each class.

\subsubsection{HLS dataset} 
To examine the proposed FCA framework on real-world distributed medical dataset, we exploit several public datasets released by different institutes for lung segmentation in X-ray images, forming a heterogeneous lung segmentation dataset (referred to as HLS). The HLS dataset consists of four independent subsets, including COVID-19 x-ray dataset\cite{covid-19-xray-dataset}, covid-chestxray dataset\cite{cohen2020covidProspective}, QaTa-COV19 dataset \cite{Qata1}, and COVID-19 Chest X-ray Segmentation dataset\cite{covid-19-chest-xray-segmentations-dataset}. Specifically, the COVID-19 X-ray dataset \cite{covid-19-xray-dataset} contains 6500 images of chest X-rays with pixel-level polygonal lung segmentation masks, among which 517 cases are from COVID-19 patients. The COVID-19 covid-chestxray dataset \cite{cohen2020covidProspective} contains 542 chest X-ray and CT images from COVID-19 patients or other viral and bacterial pneumonia (MERS, SARS, and ARDS) patients. The QaTa-COV19 dataset \cite{Qata1} consists of 9258 COVID-19 chest X-ray images collected by Qatar University and Tampere University. The COVID-19 Chest X-ray Segmentation\cite{covid-19-chest-xray-segmentations-dataset} consists of a collection of a total of 100 Chest X-ray images from the Novel Coronavirus (COVID-19) cases. We assign each subset of HLS to different clients forming multiple ``data islands" containing real heterogeneous environmental factors (such as races, surroundings, and imaging devices). The standard mDice serves as the metric to evaluate the segmentation performance on this dataset.

\subsubsection{AMD-SD-HG} 
The AMD-SD \cite{hu2024amd} dataset contains 3049 B-scan images from 138 patients with segmentation categories of subretinal effusion, subretinal effusion, elliptical continuity, subretinal hyperreflective material, and pigment epithelium detachment. Similar to the above CXRS-HG dataset, we transform it into a heterogeneous distributed dataset AMD-SD-HG with the same transformation strategies to simulate client-wise heterogeneity in practical medical scenarios. We use the standard mDice metric to evaluate the segmentation performance on this dataset.

\subsection{Implementation Details}
The threshold $\delta$ for score binarization in our BFU strategy ({Eq.}~\ref{eq5}) is empirically set as 0.25. We employ the SAM-Med2D \cite{r2} as an example of MFMs in {Fig.}~\ref{Overview}~(a) to examine our FCA framework if not specially mentioned. The layer number \(N_L\) in the image encoder ({Fig.}~\ref{Overview}~(a)) of SAM-Med2D is set as 12. As for the training details, the standard Binary Cross-Entropy (BCE) loss \cite{unet} is applied for the segmentation head, and the Cross-Entropy loss is applied to optimize the client discriminator. During federated learning, the global model in the central server is updated in total for \(T=60\) rounds. During each global round, the local clients are individually trained for \(N_e\) local epochs (\(N_e=5\) for the CXRS-HG dataset and \(N_e=3\) for the HLS dataset), then the updated local parameters will be uploaded to the central server for sequential integration, which is similar to \cite{r1}. All the models are trained via the Adam optimizer with a learning rate of 0.001 and a weight decay of 0.001 on 4 NVIDIA RTX 3090 graphics cards.

\subsection{Comparsion with State-of-the-art Methods}
To validate the effectiveness and superiority of the proposed federated client-tailored adapter (FCA) framework,we conduct comparison experiments on three large-scale heterogeneous distributed datasets for medical segmentation, including the CXRS-HG, HLS and AMD-SD-HG datasets. We report the mDice on three local clients and the average mDice across all clients for comprehensive performance comparison. 

We first compare our FCA framework with other state-of-the-art FL methods, including the FedSeg\cite{miao2023fedseg}, FedProx\cite{FedProx}, HarmoFL\cite{jiang2022harmofl}, IOP-FL \cite{jiang2023iop}, FedA3I\cite{wu2024feda3i}, FedCross\cite{xu2024federated}, IOP-FL\cite{jiang2023iop}, PerFedAvg\cite{zhou2024personalized}, and MAP\cite{li2024map}. Then, we re-implemented some MFMs enhanced variants for additional comparison. As shown in {Table}~\ref{table1}, the re-implemented MFMs enhanced variants perform significantly better than those without MFMs since stirring universal prior knowledge in MFMs helps to improve and stabilize heterogeneous federated learning (see {Fig.}~\ref{fig1}~(b)). In addition, our FCA respectively outperforms the second-best method \cite{jiang2023iop} equipped with the same MFMs by large margins of 3.51\%, 1.65\% on the  CXRS-HG, HLS datasets mainly attributed to the following reasons: (1) Unlike non-tailored FL methods (\ie, FedAvg, FedProx), our FCA allows each client to undergo customized federated updating, thus enabling the optimal tailored model for each client. (2) Although the IOP-FL \cite{jiang2023iop}, FedCross \cite{xu2024federated}, MAP\cite{li2024map}, and PerFedAvg \cite{zhou2024personalized} also attempt to customize federated models for local clients, 
they decouple at coarse-grained model or module levels \cite{10440498,shen2022cd2} and allocate parameter groups \cite{xie2024pflfe,chen2024towards} statically, lacking enough adaptability. While our FCA framework achieves better performance via dynamic and fine-grained parameter decompose. (3) Our FCA framework equipped with the smooth federated updating strategy (FCA-SFU) performs better than that equipped with the binary federated updating strategy (FCA-BFU). It is because our SFU probabilistically distributes the adapter units to multiple clients, thus achieving more fine-grained and adaptive federated updating. To further validate the generalization across different scales and modalities, we compared our FCA framework with several representative methods on the large-scale AMD-SD-HG dataset, including two baselines (FedAvg, FedAvg*\cite{r1}), a personalized approach (FedCross*\cite{xu2024federated}) , a decoupling approach (MAP*\cite{li2024map}), and the existing state-of-the-art approach (IOP-FL*\cite{jiang2023iop}). As shown in {Table}~\ref{compare_2}, our FCA framework consistently outperforms these approaches with large margins.

\begin{table}[]
\begin{center}
\caption{Experimental validation of alleviating various client-wise heterogeneity on the CXRS-HG dataset}
\begin{tabular}{l|cc}
\hline\hline
\textbf{Heterogeneity}       & \textbf{FedAvg}*\cite{r1} \textbf{(baseline\(_2\))} & \textbf{FCA-SFU (ours)} \\ \hline
IID     & 69.70           & \textbf{70.89} \\ 
Class Imbalance     & 61.78           & \textbf{64.27}                     \\
Distribution Diversity & 62.91           & \textbf{64.05}                     \\
Imbalance+Diversity  & 60.64           & \textbf{64.15}                     \\ \hline\hline

\end{tabular}
\label{table2}
\end{center}
\vspace{0cm}
\end{table}

\begin{table}[]
\begin{center}
    
\caption{Experimental validation of generalization ability for different MFMs on the CXRS-HG dataset}
\begin{tabular}{l|ll}
\hline\hline
\textbf{Medical Foundation Models (MFMs)}      & \textbf{Client}\textsubscript{k} & \textbf{Average} \\ \hline
FedAvg \cite{r1} without MFMs (baseline\(_1\))           & 35.82              & 36.94            \\ \hline
FedAvg*\cite{r1} with H-SAM\cite{cheng2024unleashing} &  59.13             & 58.36            \\
FCA-SFU with H-SAM\cite{cheng2024unleashing} (\textbf{ours}) & 62.51              & 62.15            \\
FedAvg*\cite{r1} with Med-SA\cite{zhang2024segment} &  47.10             & 45.92            \\
FCA-SFU with Med-SA\cite{zhang2024segment} (\textbf{ours}) & 53.91              & 51.53            \\
FedAvg* \cite{r1} with SAM-Med2D \cite{r2} (baseline\(_2\)) & 60.69              & 60.64            \\
FCA-SFU with SAM-Med2D\cite{r2} (\textbf{ours})          & 64.47              & 64.15            \\
\hline\hline
\end{tabular}
\vspace{0cm}

\item[*] Our implementation of corresponding MFMs enhanced variants.

\label{table3}
\end{center}
\vspace{-0.0cm}
\end{table}

In addition, we qualitatively compare the segmentation masks obtained from our FCA-SFU, the FedAvg* (the MFMs enhanced variant of the FedAvg \cite{r1}), and the original FedAvg \cite{r1} for effectiveness validation. As shown in {Fig.}~\ref{visualization} and {Fig.}~\ref{visualization2}, our FCA-SBU achieves more precision segmentation than the FedAvg* variant on every local client. Besides, the difficulty of segmenting the same anatomical structure differs across clients (\eg~the boxed areas in {Fig.}~\ref{visualization}) since the complicated client-wise heterogeneity. Conventional FL methods like the FedAvg* variant only obtain a global-compromised model for all clients that leads to a sub-optimal compromised segmentation. In contrast, our FCA-SFU customizes a local model for each client that achieves an optimal client-tailored segmentation for each client. In {Fig.}~\ref{visualization2}, some clients (such as $client_2$) require simultaneously segmenting out the shadow area that overlapped by the mediastinum and right lung (see corresponding ground-truth annotations) while other clients (\(client_1\), \(client_3\)) do not need to segment this area. Since the FedAvg* variant only has a global-compromised segmentation model, the \(client_1\) and \(client_3\) mis-segmented out this shadow area. In contrast, our FCA-SFU was not affected by this annotation heterogeneity demonstrating the superiority and effectiveness of our client-tailored adapter framework.

\subsection{Ablation Studies}
To examine the effectiveness of each component, we compare our FCA with a series of baselines and variants, the main results already contained in {Table}~\ref{table1}. Specifically, the FCA-BFU and FCA-SFU  respectively are the proposed Federated Client-tailored Adapter (FCA) equipped with our binary (BFU) and smooth (SFU) federated updating strategies. FedAvg* \cite{r1} (baseline\(_2\)) is a baseline constructed with the same network (\ie~SAM-Med2D) as our FCA except it utilizes a conventional federated updating strategy FedAvg \cite{r1}. The FedAvg\cite{r1} (baseline\(_1\)) is the vanilla  FedAvg baseline without assistance from off-the-shelf MFMs. The results of baseline\(_1\) and baseline\(_2\) indicate that the prior knowledge contained in off-the-shelf MFMs indeed benefits a lot for improving segmentation performance ({Table}~\ref{table1}) and stabilizing the heterogeneous federated updating (see {Fig.}~\ref{fig1}~(b)). Eventually, our client-tailored adapter FCA-SFU further improves the individual performance of each client and sets a new start-of-art result on both CXRS-HG and HLS datasets.

\subsection{Experimental Validation of Alleviating Heterogeneity}
In this section, we further examine the capability of our FCA for alleviating various client-wise heterogeneity and providing client-tailored segmentation. Since the CXRS-HG dataset contains two kinds of heterogeneity including class imbalance and distribution diversity, we split them alone and construct corresponding dataset variants to train two variant models. As shown in {Table}~\ref{table2}, our FCA-SFU significantly boosts performance (mDice) from 61.78 to 64.27 and from 62.91 to 64.05 regarding heterogeneity types of class imbalance and distribution diversity. When dealing with the more complicated heterogeneity mixture (Imbalance+Diversity), the performance drop from our FCA-SFU is negligible while the drop from FedAvg*\cite{r1} is prominent. This difference further validates that our FCA-SFU is good at learning client-tailored models for each client and effectively alleviating various heterogeneity existing in distributed medical image segmentation.

\begin{table}[t]
\begin{center}

\caption{Comparison and analysis of computational efficiency on the CXRS-HG dataset}
\begin{tabular}{l|cc}
\hline
\hline
          Methods& \#Params (M)  & \#Time (h) \\ \hline
FedAvg* (baseline\(_2\))   & 180.50 & 36.14  \\
FedCross* & 180.50 & 108.39  \\
IOP-FL*   & 180.50 & 39.61  \\
FCA-SFU* (\textbf{ours})  & 180.59 & 36.64  \\ \hline\hline
\end{tabular}
\vspace{0.1cm}

\item[*] Our implementation of SAM-Med2D enhanced variants.

\label{table4}
\end{center}
\end{table}

\subsection{ Experimental Validation of MFMs Generalization}
In this section,  we validate that our federated client-tailored adapter framework is generalizable to various MFMs consisting of transformers. As shown in {Table}~\ref{table3}, we examine our FCA-SFU on three medical foundation models including the SAM-Med2D \cite{r2}, Med-SA~\cite{wu2023medical} and H-SAM\cite{cheng2024unleashing}. We observe that our FCA-SFU shows consistent improvements over the conventional FL baseline method FedAvg*\cite{r1}, indicating the generalization ability of the proposed FCA framework for various MFMs.

\subsection{Comparison and Analysis of Computational Efficiency}
In this section, we compare and analyze the computational efficiency by measuring the model parameters (\#Params) and convergence time (\#Time), respectively. For fair comparisons, all the methods are implemented and enhanced by the same MFMs (SAM-Med2D\cite{r2}). Besides, since all compared methods nearly involve the same parameter transmission and collection process, the model parameters (\#Params) for each local model copy could also be treated as an indirect metric to evaluate the communication overhead during federated learning. As shown in {Table}~\ref{table4}, our FCA-SFU achieves significant performance improvement with a slight increase in model parameters and communication overhead. Moreover, the convergence speed of our FCA-SFU is much faster than other state-of-the-art methods.

\section{Conclusion}
This paper introduces a generalizable framework Federated Adaptive Adapter (FCA) for heterogeneous medical image segmentation. We identify the training instability issue induced by client-wise heterogeneity in the conventional federated learning paradigm and propose two measures to alleviate it. One measure distills the universal knowledge in off-the-shelf medical foundation models to stabilize heterogeneous federated learning via the parameter-efficient adapter. Another measure decomposes the adapter units in each client into client-specific and client-invariant parts and develops different federated updating strategies for them. This measure further stabilizes the heterogeneous federated training process and realizes client-tailor federated learning at the same time. Eventually, the FCA achieves state-of-the-art performance on three datasets for heterogeneous medical image segmentation.

\bibliographystyle{ieeetr}
\bibliography{ref}

\end{document}